\documentclass[runningheads]{llncs}
\usepackage[T1]{fontenc}
\usepackage{graphicx}
\usepackage{booktabs}
\usepackage{amsmath}
\usepackage{hyperref}
\usepackage{cite}
\usepackage{amsmath,amssymb,amsfonts}
\usepackage{algorithmic}
\usepackage{graphicx}
\usepackage{textcomp}
\usepackage{multirow}
\usepackage{booktabs}
\usepackage{makecell}
\usepackage{xcolor}
\usepackage{float}

\begin{document}
%
\title{Boosting Medical Image Synthesis via Registration-guided Consistency and Disentanglement Learning}
\titlerunning{Registration-guided Consistency and Disentanglement Learning}
%
\author{Chuanpu Li\inst{1, 2, 3}, Zeli Chen\inst{1, 2, 3}\thanks{Chuanpu Li and Zeli Chen contribute equally to this article.}, Yiwen Zhang\inst{1, 2, 3}, Liming Zhong\inst{1, 2, 3}, and Wei Yang\inst{1, 2, 3}\thanks{Corresponding author: Wei Yang (weiyanggm@gmail.com).}}
\institute{School of Biomedical Engineering, Southern Medical University, Guangzhou, China \and Guangdong Provincial Key Laboratory of Medical Image Processing, Southern Medical University, Guangzhou, China \and Guangdong Province Engineering Laboratory for Medical Imaging and Diagnostic Technology, Southern Medical University, Guangzhou, China}
\maketitle              

\begin{abstract}
    Medical image synthesis remains challenging due to misalignment noise during training. Existing methods have attempted to address this challenge by incorporating a registration-guided module. However, these methods tend to overlook the task-specific constraints on the synthetic and registration modules, which may cause the synthetic module to still generate spatially aligned images with misaligned target images during training, regardless of the registration module's function. Therefore, this paper proposes registration-guided consistency and incorporates disentanglement learning for medical image synthesis. The proposed registration-guided consistency architecture fosters task-specificity within the synthetic and registration modules by applying identical deformation fields before and after synthesis, while enforcing output consistency through an alignment loss. Moreover, the synthetic module is designed to possess the capability of disentangling anatomical structures and specific styles across various modalities. An anatomy consistency loss is introduced to further compel the synthetic module to preserve geometrical integrity within latent spaces. Experiments conducted on both an in-house abdominal CECT-CT dataset and a publicly available pelvic MR-CT dataset have demonstrated the superiority of the proposed method. 
\end{abstract}
\section{Introduction}
\begin{figure}[t]
    \includegraphics[width=\textwidth]{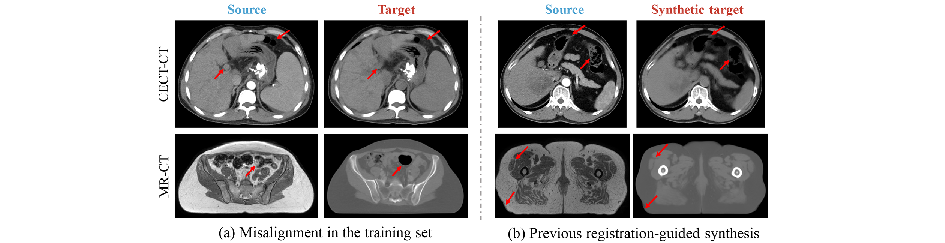}
    \caption{The misaligned challenges in medical image synthesis. (a) Misaligned anatomical structures result in inaccurate mapping in supervised paradigms. (b) The previous registration-guided synthesis \cite{kong2021breaking} still retains misalignment-induced noise and suffers from blurred boundaries due to the averaging effect.}
\label{challenges}
\end{figure}
Medical image synthesis can offer advantages by predicting a target imaging modality in situations where acquisition is impractical due to time and cost constraints, or when image registration introduces unacceptable uncertainty between images of different modalities \cite{zhang2023generalized, armanious2020medgan}. For example, MR-to-CT synthesis is a clinically significant solution for MRI-only radiation therapy treatment planning \cite{dayarathna2023deep, chen2023multi}, and CECT-to-CT synthesis is beneficial for reducing the registration errors between contrast-enhanced CT (CECT) and CT, the additional exposure, and treatment cost \cite{sumida2019deep, pang2023ncct}.

Existing medical image synthesis methods can be categorized into supervised and unsupervised learning paradigms. In the supervised setting, source and target modalities are registered prior to training \cite{chen2023multi, dalmaz2022resvit, pang2023ncct}. However, as shown in Fig. \ref{challenges}(a), the standard image pre-processing steps cannot entirely eliminate misaligned errors, particularly in anatomical areas like the abdomen or pelvis. This misalignment can result in supervised methods learning inaccurate mappings \cite{chen2023multi, dalmaz2022resvit, zhou2023mitigating, zhong2023united}. Unsupervised methods alleviate the necessity of paired images. Unfortunately, these methods cannot fully utilize retrospective images from the same patient, and the performance of unsupervised synthesis methods is inferior to supervised ones \cite{phan2023structure, dayarathna2023deep, zhong2023qacl}. Therefore, better image quality may be obtained by trying to solve the misalignment problem in the supervised paradigm.

Recent methods impose the registration-guided module to mitigate the misalignment challenge \cite{zhong2023qacl, zhong2023united, kong2021breaking, zhou2023mitigating, honkamaa2023deformation, su2023angiomoco}. For example, Kong et al. proposed RegGAN to train the generator using an additional registration network to fit the misaligned noise distribution adaptively for T1-to-T2 MR image synthesis\cite{kong2021breaking}. Zhong et al. proposed quartet attention aware closed-loop learning (QACL) framework for MR-to-CT synthesis via simultaneous registration\cite{zhong2023qacl}. However, these methods neglect the task-specific constraints on synthetic and registration modules. Specifically, the synthetic module should ideally preserve geometric properties, while the registration module should effectively learn geometric transformations from images suffering from misalignment. \textit{This neglect could potentially cause the synthetic module to directly generate spatially aligned images with the misaligned target image during training, regardless of the function of the registration block} \cite{arar2020unsupervised}. In the testing phase, images synthesized solely using the synthetic network may still retain misalignment-induced noise and suffer from blurred boundaries due to the averaging effect. (As shown in Fig. \ref{challenges}(b)).

Some methods also integrate disentanglement learning to address the challenge of misalignment \cite{lin2022deep, qin2019unsupervised}. These methods disentangle the anatomical structures and misalignment into distinct features in the latent spaces and synthesize target images using only anatomical features. However, \textit{different styles between modalities may also be falsely interpreted as motions, resulting in implausible deformations} \cite{wang2021dicyc, reaungamornrat2022multimodal}. Therefore, distinguishing the anatomy and the styles from different modalities and constraining anatomical features to be consistent is also beneficial for alleviating the misalignment problem.

In this study, we boost medical image synthesis via registration-guided consistency and disentanglement learning. Unlike prior methods which neglect the task-specific constraints in both synthetic and registration modules, our proposed approach integrates a registration-guided consistency architecture alongside an alignment loss, aiming at minimizing the difference between the images synthesized before and after applying the same deformation field. Through that, the synthetic network is encouraged to eliminate the influence of misalignment-induced noise, under the presumption of consistent output across diverse deformed inputs. Furthermore, we propose an anatomy consistency disentanglement synthetic module (ACDS) to decompose anatomical and style features. The disentangled style features prove advantageous for the registration network in discerning authentic deformations \cite{reaungamornrat2022multimodal}. For the disentangled anatomical features, we introduce an anatomy consistency loss to further mitigate the misalignment issue within the latent space. Experimental validations conducted on both an in-house abdominal CECT-CT dataset and a publicly available pelvic MR-CT dataset demonstrate our method underscores the efficacy relative to several synthetic approaches.

\section{Method}

\begin{figure}[t]
    \centering
    \includegraphics[width=\textwidth]{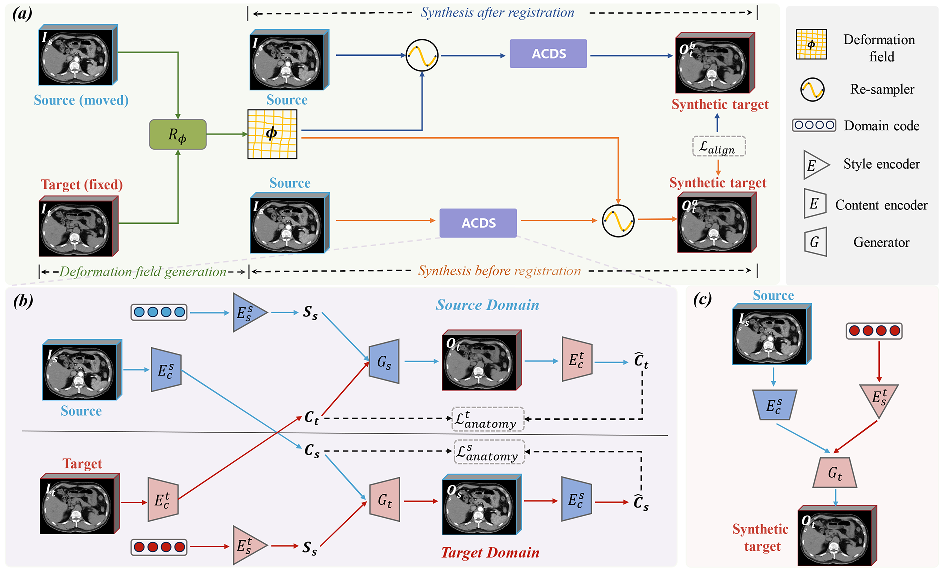}
    \caption{Overview of our proposed methods. (a) The registration-guided consistency architecture. $\mathcal{L}_{align}$ is used to minimize the synthetic images from synthesis after registration and synthesis before registration. (b) The anatomy consistency disentanglement synthetic (ACDS) module. (c) Testing phase for medical image synthesis.}
\label{framework}
\end{figure}

Let $ \mathcal{S} \subset \mathbb{R}^{D_S \times H_S \times W_S} $ and $ \mathcal{T} \subset \mathbb{R}^{D_T \times H_T \times W_T} $ represent domains of source image and target image, where $D, H, W$ are the slice numbers, height, and weight. We denote $ \mathbf{S}={\{I_{s} \in \mathcal{S}\}}_{s=1}^{N} $ and $ \mathbf{T}={\{I_{t} \in \mathcal{T}\}}_{t=1}^{N} $ as paired image sets. Medical image synthesis aims to learn a synthetic network $G$ that maps the source image $I_{s}$ to its target image $O_{t} = G(I_{s})$. However, the inevitable misalignment between $\mathbf{S}$ and $\mathbf{T}$ often results in inaccurate mapping. To address this challenge, we propose a registration-guided consistency disentanglement learning method. As shown in Fig. \ref{framework}, the proposed scheme consists of two major parts: (a) a registration-guided consistency architecture, (b) an anatomy consistency disentanglement synthetic (ACDS) module. Below, we describe the design and objective of each part.

\subsection{Registration-guided Consistency}

A key challenge of registration-guided medical synthesis lies in delineating the respective roles of the synthetic network and registration network. Specifically, it's crucial to avoid situations where the synthetic network ignores the function of the registration network and directly synthesizes spatially aligned images during training. To achieve this, we assume that the synthetic and registration processes are commutative and propose a registration-guided consistency structure (As shown in Fig. \ref{framework}(a)).

\subsubsection{Registration-guided Module.} The registration module is composed of a deformation generator $R_{\Phi}$ and a re-sampler $R_{S}$. $R_{\Phi}$ is designed to take $I_{s}$ and $I_{t}$ as input, and produce a deformation field $\phi$ describing the non-rigid alignment from $I_{s}$ to $I_{t}$: 
\begin{equation} 
    \phi = R_{\Phi}\left(I_s, I_t\right).
\end{equation}
$R_{S}$ receives the deformation field $\phi$ and applies it to an image desired to be moved. We have adopted the architectural framework detailed in \cite{balakrishnan2019voxelmorph} as the foundation of our registration module, and incorporated its smoothing loss to preserve the smooth displacement of the deformable field:

\begin{equation} 
    \mathcal{L}_{\text {smooth }}(\phi)=\mathbb{E}_{\mathcal{S}, \mathcal{T}}[\|\nabla {\mathbf{\phi}}\|^2].
\end{equation}

\subsubsection{Synthesis Before Registration.} Synthesis before registration first applies the synthetic module $G$ on $I_{s}$, followed by the utilization of the re-sampler layer $R_{S}$ on the synthesized image. This process yields the final output, which can be formally expressed as: 

\begin{equation}   
    O_{t}^{b}=R_{S}\left(G\left(I_s\right), \phi\right).
    \label{eq_obt}
\end{equation}

\subsubsection{Synthesis After Registration.} In contrast, synthesis after registration first applies the re-sampler layer $R_{S}$ on $I_{s}$, followed by the utilization of the synthetic module $G$ on the warped image. This process yields the other final output, which can be formally expressed as:

\begin{equation}   
    O_{t}^{a}=G\left(R_{S}\left(I_s, \phi\right)\right).
    \label{eq_oat}
\end{equation}

\subsubsection{Alignment Loss.} In Eq.\ref{eq_obt} and Eq.\ref{eq_oat}, the deformation field $\phi$ used in the re-sampler $R_{S}$ is both given by the $R_{\Phi}(I_s, I_t)$ and designed to be the same. The sole difference is that the synthetic module $G$ accepts geometrically different images as inputs ($I_{s}$ and $R_{S}(I_s, \phi)$). Thus, if we assume that the synthetic and registration processes are commutative, which expect the same output from synthesis before and after registration, the synthetic module $G$ is required to be geometry preserving and finally alleviate the misalignment issue. This assumption can be easily realized by a simple alignment loss, which can be formulated as:
\begin{equation}   
    \mathcal{L}_{\text {align}}(G, R)=\left\|O_{t}^{b}-I_t\right\|_1+\left\|O_{t}^{a}-I_t\right\|_1.
\end{equation}

\subsection{Anatomy Consistency Disentanglement Synthetic Module}

Given that variations in styles across modalities may be falsely interpreted as motions by the registration network \cite{lin2022deep}, it is beneficial for the synthetic module $G$ to disentangle the specific style for each modality. Moreover, disentangling anatomical structures and enforcing their consistency across identical content images is important in mitigating misaligned noise. Therefore, we propose an anatomy consistency disentanglement synthetic module.

\subsubsection{Synthetic process.} As shown in Fig. \ref{framework}(b), the ACDS module is composed of two content encoders $\{E_{c}^{s}, E_{c}^{t}\}$, two style encoders $\{E_{s}^{s}, E_{s}^{t}\}$, and two generators $\{G_{s}, G_{t}\}$. The encoders in both domains respectively encode an input image to an anatomical-invariant space and a modality-specific style space, and the generators respectively combine the anatomical space features and style features to synthesize images. This can be formulated as:
\begin{equation}   
    {O}_{t} = G_{t}(c_{s}, s_{t}) = G_{t}(E_{c}^{s}(I_{s}), E_{s}^{t}(d_{t})), 
\end{equation}
\begin{equation}   
    {O}_{s} = G_{s}(c_{t}, s_{s}) = G_{s}(E_{c}^{t}(I_{t}), E_{s}^{s}(d_{s})),
\end{equation}
where ${O}_{t}$ and ${O}_{s}$ denote the synthetic images, $d_{t}$ and $d_{s}$ denote the learned domain code, which is empirically set as an 8-bit vector. 

\subsubsection{Disentanglement Learning.} Inspired by \cite{liu2017unsupervised, huang2018multimodal}, the synthetic module realizes disentanglement via generative adversarial loss, self-reconstruction loss, and cycle-consistency loss. We define them as follows:

\begin{equation}
    \begin{aligned}
        \mathcal{L}_{\text{adv}} & = \mathbb{E}_{\mathcal{T}}[\log D_{t}(I_{t})] + \mathbb{E}_{\mathcal{S}}[1 -\log D_{t}({O}_{t})] \\ 
       & + \mathbb{E}_{\mathcal{S}}[\log D_{s}(I_{s})] + \mathbb{E}_{\mathcal{T}}[1 -\log D_{s}({O}_{s})],
    \end{aligned}
\end{equation}

\begin{equation}   
    \mathcal{L}_{\text {self }}=\left\|G_{s}\left(E_{c}^{s}(I_s), E_{s}^{s}(d_{s})\right)-I_{s}\right\|_1+\left\|G_{t}\left(E_{c}^{t}(I_{t}), E_{s}^{t}(d_{t})\right)-I_{t}\right\|_1, 
\end{equation}
\begin{equation}   
    \mathcal{L}_{\text {cycle}}=\left\|G_{s}\left(E_{c}^{t}({O}_{t}), E_{s}^{s}(d_{s})\right)-I_{s}\right\|_1+\left\|G_{t}\left(E_{c}^{s}({O}_{s}), E_{s}^{t}(d_{t})\right)-I_{t}\right\|_1, 
\end{equation}
where $D_{t}$ and $D_{s}$ represent the discriminators. If the encoders can successfully disentangle the anatomical content from images and the generators uphold geometric fidelity, anatomical content features should be identical. Therefore, we propose an anatomy consistency loss:
\begin{equation}   
    \mathcal{L}_{\text {anatomy}} = \left\|E_{c}^{t}({O}_{t})-E_{c}^{s}(I_{s})\right\|_1+\left\|E_{c}^{s}({O}_{s})-E_{c}^{t}(I_{t})\right\|_1. 
\end{equation}
The implementation of the anatomy consistency loss necessitates the ACDS module to further learn and retain anatomical structure information in the latent space. Then, the overall loss function is:
\begin{equation}   
    \mathcal{L}_{\text {total}} = \mathcal{L}_{\text {adv}} + \mathcal{L}_{\text {self}} + \mathcal{L}_{\text {cycle}} + \lambda_{1} \mathcal{L}_{\text {anatomy}} + \lambda_{2}\mathcal{L}_{\text {smooth}} + \lambda_{3}\mathcal{L}_{\text {align}},
\end{equation}
where $\lambda$ is the hyperparameter. As shown in Fig. \ref{framework}(c), after training, only $E_{c}^{s}$, $E_{s}^{t}$ and $G_{t}$ are required to translate the source image into target image in the testing phase. The content encoder consists of 3 Conv-IN-ReLU blocks with kernel sizes of 3, strides of 2, channels of 32, 64, 128, followed by four residual blocks. The style encoder consists of 3 MLP layers.

\section{Experiments}
\subsection{Experimental Setup.}

\subsubsection{Datasets.} We evaluated our methods on an in-house abdominal CECT-CT dataset and a publicly available pelvic MR-CT derived from the SynthRAD2023 Grand Challenge dataset \cite{thummerer2023synthrad2023}. The CECT-CT dataset comprised 278 patients diagnosed with liver cancer. Preprocessing of images involved sequential implementation of affine spatial normalization \cite{jenkinson2012fsl} and nonlinear registration techniques \cite{avants2009advanced}, aimed at aligning the CECT images with the corresponding CT images. For evaluation, we manually selected 15 well-aligned subjects for the validation set and 40 well-aligned subjects for the test set. The pelvic MR-CT dataset consists of 180 patients. Similarly, the validation set comprised 6 aligned subjects, while the test set included 30 appropriately aligned subjects. 

\subsubsection{Implementation Details} Our method was implemented with PyTorch, using 2 GPU of NVIDIA RTX3090 GPUs. The hyperparameter $\lambda_{1}$, $\lambda_{2}$, $\lambda_{3}$ were set as $0.5$, $10$ and $20$. For the anatomy consistency disentanglement synthetic module, we used a similar structure as \cite{huang2018multimodal} and changed it into 3D. We employed Adam optimizer with a poly decay strategy, a batch size of 2, and a maximum of 200 epochs to dynamically adjust the learning rate from $ 2 \times 10^{-4}$.

\subsubsection{Evaluation Metrics.} To evaluate the efficacy of our proposed model, we employed three standard evaluation metrics: the mean absolute error (MAE), the structural similarity index measurement (SSIM), and the peak signal-to-noise ratio (PSNR). Calculations for MAE, SSIM, and PSNR were confined within the body mask, with SSIM parameters set under the guidelines established by Wang et al. \cite{wang2004image}. 

\subsection{Results}

\subsubsection{Comparisons with State-of-the-Art.} We compared our method with several synthesis methods including supervised methods such as Pix2pix\cite{pix2pix2017}, ResViT\cite{dalmaz2022resvit}, unsupervised methods such as CycleGAN\cite{zhu2017unpaired}, disentanglement methods such as UNIT\cite{liu2017unsupervised}, and registration-guided method such as RegGAN\cite{kong2021breaking}.

\begin{table}
    \caption{Quantitative comparison of different methods on the CECT-to-CT synthesis and MR-to-CT synthesis.}
    \renewcommand\arraystretch{1.2}
    \resizebox{\textwidth}{!}{
    \begin{tabular}{l|l|l|l|l|l|l}
    \hline
    Methods & \multicolumn{3}{c|}{CECT-to-CT Synthesis} & \multicolumn{3}{c}{MR-to-CT Synthesis} \\ \cline{2-7} 
    & MAE (HU) $\downarrow$ & PSNR (dB) $\uparrow$ & SSIM (\%) $\uparrow$ & MAE (HU) $\downarrow$ & PSNR (dB) $\uparrow$ & SSIM (\%) $\uparrow$ \\ \hline
    Pix2pix\cite{pix2pix2017} & 21.70 $\pm$ 1.40 & 34.89 $\pm$ 1.43 & 87.78 $\pm$ 3.02 & 50.64 $\pm$ 4.47 & 30.61 $\pm$ 0.85 & 82.82 $\pm$ 2.14 \\
    CycleGAN\cite{zhu2017unpaired} & 22.51 $\pm$ 1.78 & 34.33 $\pm$ 1.12 & 88.14 $\pm$ 2.78 & 82.24 $\pm$ 11.42 & 27.45 $\pm$ 0.79 & 70.79 $\pm$ 6.51 \\
    UNIT\cite{liu2017unsupervised} & 22.78 $\pm$ 1.80 & 34.04 $\pm$ 1.06 & 88.10 $\pm$ 2.78 & 84.49 $\pm$ 11.95 & 27.80 $\pm$ 0.88 & 70.44 $\pm$ 6.22 \\
    RegGAN\cite{kong2021breaking} & 21.29 $\pm$ 1.86 & 35.10 $\pm$ 1.27 & 88.15 $\pm$ 2.84 & 50.22 $\pm$ 3.60 & 30.44 $\pm$ 0.53 & 83.16 $\pm$ 1.92 \\
    ResViT\cite{dalmaz2022resvit} & 20.79 $\pm$ 2.09 & 35.14 $\pm$ 1.41 & 88.20 $\pm$ 2.89 & 49.90 $\pm$ 3.43 & 30.72 $\pm$ 0.60 & 82.91 $\pm$ 2.07 \\
    \textbf{Ours} & \textbf{16.38 $\pm$ 1.83} & \textbf{36.52 $\pm$ 1.66} & \textbf{89.22 $\pm$ 2.83} & \textbf{47.47 $\pm$ 4.59} & \textbf{30.86 $\pm$ 0.88} & \textbf{83.70 $\pm$ 2.47} \\
    \hline
    \end{tabular}}
    \label{tab:compare}
\end{table}

\begin{figure}
    \vspace{-0.5cm}
    \centering
    \includegraphics[width=\textwidth]{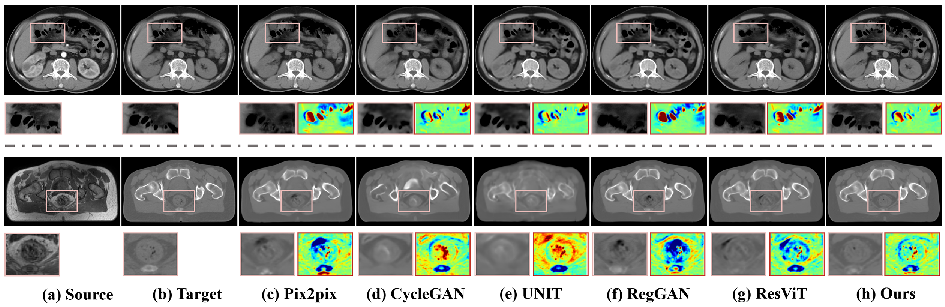}
    \caption{Visual comparison of synthesized results on CECT-CT and MR-CT datasets. The pink boxes show the structural details where misalignment are often encountered during training.}
\label{results}
\end{figure}

As illustrated in Table \ref{tab:compare}, our proposed method demonstrates superior performance, exhibiting a MAE of 16.38 $\pm$ 1.83, a PSNR of 36.52 $\pm$ 1.66 and an SSIM of 89.22 $\pm$ 2.83 for CECT-to-CT synthesis, and an MAE of 47.47 $\pm$ 4.59, a PSNR of 30.86 $\pm$ 0.88 and a SSIM of 83.70 $\pm$ 2.47 for MR-to-CT synthesis. Fig. \ref{results} shows the visual results. The supervised and previous registration-guided methods still retain misalignment noise and have blurred boundaries due to the averaging effect. In contrast, our methods preserve anatomical consistency and clear boundaries in the regions where misalignment is often encountered during training.

\subsubsection{Ablation Study.} An ablation study was undertaken to assess the efficacy of the registration-guided consistency and ACDS module. Table \ref{tab:ablation} shows the results obtained from the validation set for both CECT-to-CT and MR-to-CT tasks. The third row in Table \ref{tab:ablation} demonstrates that the registration-guided consistency design yields a reduction of 2-3 HU in the MAE compared with synthesis before registration in both tasks. The results in the fourth row also indicate the integration of the ACDS module facilitates an enhancement in the quality of synthesis.
\begin{table}
    \caption{Ablation study of critical components of the proposed method. AFT represents synthesis after registration, BEF represents synthesis after registration, and ACDS represents anatomy consistency disentanglement synthetic module.}
    \renewcommand\arraystretch{1.2}
    \resizebox{\textwidth}{!}{
    \begin{tabular}{c|c|c|l|l|l|l|l|l}
    \hline
    \multicolumn{3}{l|}{Component} & \multicolumn{3}{c|}{CECT-to-CT Synthesis} & \multicolumn{3}{c}{MR-to-CT Synthesis} \\ \cline{1-9} 
    AFT & BEF & ACDS & MAE (HU) $\downarrow$ & PSNR (dB) $\uparrow$ & SSIM (\%) $\uparrow$ & MAE (HU) $\downarrow$ & PSNR (dB) $\uparrow$ & SSIM (\%) $\uparrow$ \\ 
    \hline
    \textbullet & \textopenbullet & \textopenbullet & 28.84 $\pm$ 2.70 & 33.49 $\pm$ 1.07 & 85.30 $\pm$ 3.06 & 86.84 $\pm$ 11.23 & 27.61 $\pm$ 1.01 & 69.74 $\pm$ 2.30 \\
    \textopenbullet & \textbullet & \textopenbullet & 21.56 $\pm$ 1.49 & 34.65 $\pm$ 1.03 & 87.22 $\pm$ 3.11 & 55.59 $\pm$ 3.33 & 29.65 $\pm$ 0.31 & 81.74 $\pm$ 1.59 \\
    \textbullet & \textbullet & \textopenbullet & 19.43 $\pm$ 1.84 & 35.30 $\pm$ 1.09 & 87.47 $\pm$ 3.12 & 53.48 $\pm$ 6.56 & 30.12 $\pm$ 0.65 & 81.98 $\pm$ 1.52 \\
    \textbullet & \textbullet & \textbullet & \textbf{16.96 $\pm$ 1.92} & \textbf{35.64 $\pm$ 1.34} & \textbf{88.27 $\pm$ 3.16} & \textbf{52.68 $\pm$ 5.01} & \textbf{30.18 $\pm$ 0.35} & \textbf{82.20 $\pm$ 1.40} \\
    \hline
    \end{tabular}}
    \label{tab:ablation}
\end{table}

\section{Conclusion}

This paper introduces a registration-guided consistency architecture and an anatomy consistency disentanglement synthetic module to address the misalignment issue for medical image synthesis. The proposed registration-guided consistency design customizes the registration and synthetic module to be task-specific and encourages the synthetic module to eliminate the influence of misalignment-induced noise. Additionally, disentanglement learning and an anatomy consistency loss are applied to the synthetic module, enhancing its ability to preserve geometric integrity, thus further avoiding the misalignment issue. Experimental results conducted on both an in-house CECT-CT dataset and a publicly available MR-CT dataset have demonstrated the superiority of the proposed method.

%
%
\bibliographystyle{splncs04}
\bibliography{mybibliography}
%




\end{document}